\documentclass[10pt,twocolumn,letterpaper]{article}

\usepackage{cvpr}
\usepackage{times}
\usepackage{epsfig}
\usepackage{graphicx}
\usepackage{amsmath}
\usepackage{amssymb}
\usepackage{soul}
\usepackage{multirow}
\usepackage{csquotes}
\usepackage{enumerate}
\usepackage[normalem]{ulem}
\usepackage{algorithm}
\usepackage{algorithmicx}
\usepackage{algpseudocode}
\usepackage{booktabs}
\usepackage{array}
\usepackage{overpic}
\usepackage{rotating}

\usepackage[pagebackref=false,breaklinks=true,letterpaper=true,colorlinks,bookmarks=false]{hyperref}

\cvprfinalcopy 

\def\cvprPaperID{2097} 

\def\ourmodel{\textit{UC-Net}}
\ifcvprfinal\fi

\graphicspath{{./Imgs/}}
\DeclareGraphicsExtensions{.jpg,.pdf,.png}
\begin{document}

\title{UC-Net: Uncertainty Inspired RGB-D Saliency Detection \\via Conditional Variational Autoencoders
}

\author{
Jing Zhang$^{1,4,5}$\quad
Deng-Ping Fan$^{2,6,}$\thanks{Corresponding author: Deng-Ping Fan \emph{(dengpfan@gmail.com)}}\quad
Yuchao Dai$^{^3}$\quad
Saeed Anwar$^{1,5}$\\
Fatemeh Sadat Saleh$^{1,4}$\quad
Tong Zhang$^1$\quad
Nick Barnes$^1$\\
$^1$ Australian National University \quad
$^2$ CS, Nankai University \quad
$^3$ Northwestern Polytechnical University \\
$^4$ ACRV \quad
$^5$ Data61 \quad 
$^6$ Inception Institute of Artificial Intelligence (IIAI), Abu Dhabi, UAE\\
}


\def\JZ#1{{\color{red}{\bf [Jing:} {\it{#1}}{\bf ]}}}
\def\YD#1{{\color{blue}{\bf [Yuchao:} {\it{#1}}{\bf ]}}}
\def\SA#1{{\color{red}{\bf [Saeed:} {\it{#1}}{\bf ]}}}
\def\NB#1{{\color{green}{\bf [Nick:} {\it{#1}}{\bf ]}}}
\newcommand{\FS}[1]{\textcolor{magenta}{{\bf #1}}}
\newcommand{\fs}[1]{\textcolor{magenta}{ #1}}
\newcommand{\fdp}[1]{{\textcolor{red}{#1}}}
\def\TZ#1{{\color{cyan}{\bf [Tong:} {\it{#1}}{\bf ]}}}
\maketitle
\thispagestyle{empty}

\begin{abstract}

In this paper, we propose the first framework (\textbf{\ourmodel})~to employ uncertainty for RGB-D saliency detection by learning from the data labeling process. Existing RGB-D saliency detection methods treat the saliency detection task as a point estimation problem, and produce a single saliency map following a deterministic learning pipeline. 
Inspired by the saliency data labeling process, we propose probabilistic RGB-D saliency detection network via conditional variational autoencoders to model human annotation uncertainty
and generate multiple saliency maps for each input image by sampling in the latent space. 
With the proposed saliency consensus process, we are able to generate an accurate saliency map based on these multiple predictions. 
Quantitative and qualitative evaluations on six challenging benchmark datasets against 18 competing algorithms demonstrate the effectiveness of our approach in learning the distribution of saliency maps, leading to a new state-of-the-art in RGB-D saliency detection\footnote{Our code is publicly available at: \url{https://github.com/JingZhang617/UCNet}.}.

\end{abstract}

\section{Introduction}
Object-level visual saliency detection involves separating the most conspicuous objects that attract humans from the background \cite{itti_saliency,achanta2009frequency,Iter_Coop_CVPR,Zhang_2018_CVPR,Liu_2019_ICCV,F3Net_aaai2020,jing2020weakly}. Recently, visual saliency detection from RGB-D images have attracted lots of interest due to the importance of depth information in human vision system and the popularity of depth sensing technologies \cite{dmra_iccv19,zhao2019Contrast}. Given a pair of RGB-D images, the task of RGB-D saliency detection aims to predict a saliency map by exploring the complementary information between color image and depth data. 


The de-facto standard for RGB-D saliency detection is to train a deep neural network using ground truth (GT) saliency maps provided by the corresponding benchmark datasets, where the GT saliency maps are obtained through human consensus or by the dataset creators \cite{sip_dataset}. 
Building upon large scale RGB-D datasets, deep convolutional neural network based models \cite{Fu2020JLDCF,dmra_iccv19,chen2019three,han2017cnns} have made profound progress in learning the mapping from an RGB-D image pair to the corresponding GT saliency map. Considering the progress for RGB-D saliency detection under this pipeline, in this paper, we would like to argue that this pipeline fails to capture the \textit{uncertainty} in labeling the GT saliency maps. 

\begin{figure}[t!]
	\centering
    \small
	\begin{overpic}[width=.98\columnwidth]{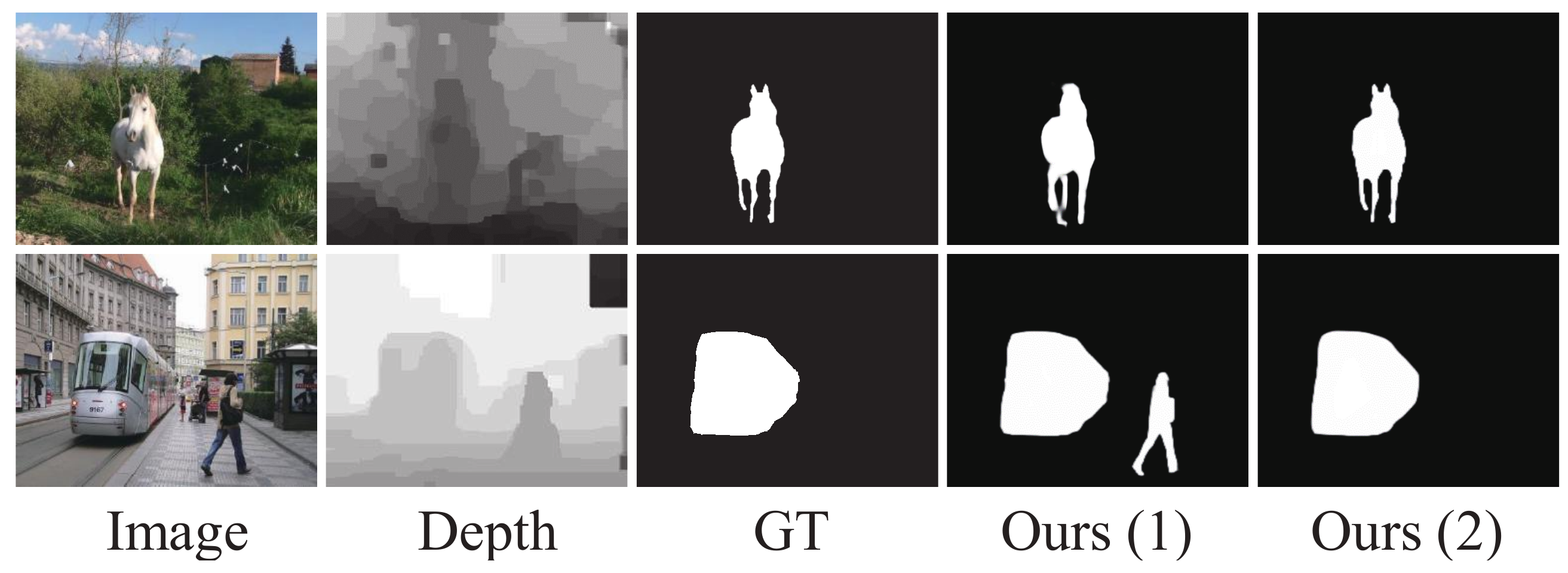}
    \end{overpic}
    \caption{Provided GT compared with UC-Net (ours) predicted saliency maps. For images with a single salient object (1 $^{st}$ row), we can produce consistent prediction. When multiple salient objects exist (2$^{nd}$ row),
   we can produce diverse predictions.
   }
    \label{fig:inconsistent_ef_sod}
    \vspace{-4mm}
\end{figure}

According to research in human visual perception \cite{scanpath}, visual saliency detection is subjective to some extent. Each person could have specific preferences in labeling the saliency map (which has been previous discussed in user-specific saliency detection \cite{ITTI20001489}). Existing approaches to RGB-D saliency detection treat saliency detection as a point estimation problem, and produce a single saliency map for each input image pair following a \textit{deterministic} learning pipeline, which fails to capture the stochastic characteristic of saliency, and may lead to a partisan saliency model as shown in second row of Fig. \ref{fig:inconsistent_ef_sod}. 
Instead of obtaining only a single saliency prediction (point estimation), we are interested in how the network produces multiple predictions (distribution estimation), which are then processed further to generate a single prediction in a similar way to how the GT saliency maps are created.

In this paper, inspired by human perceptual uncertainty, we propose a conditional variational autoencoders \cite{structure_output} (CVAE) based RGB-D saliency detection model \ourmodel~to produce multiple saliency predictions by modeling the distribution of output space as a generative model conditioned on the input RGB-D images to account for the human uncertainty in annotation.

However, there still exists one obstacle before we could apply the probabilistic framework, that is existing RGB-D benchmark datasets generally only provide a single GT saliency map for each RGB-D image pair. 
To produce diverse and accurate predictions\footnote{Diversity of prediction is related to the content of image. Image with clear content may lead to consistent prediction (1$^{st}$ row in Fig. \ref{fig:inconsistent_ef_sod}), while complex image may produce diverse predictions (2$^{nd}$ row of Fig. \ref{fig:inconsistent_ef_sod}).}, we resort to the ``hide and seek'' \cite{hide_and_seek-iccv2017} principle following the orientation shifting theory \cite{ITTI20001489} by iteratively hiding the salient foreground from the RGB image for testing, which forces the deep network to learn the saliency map with diversity. 
Through this iterative hiding strategy, we obtain multiple saliency maps for each input RGB-D image pair, which reflects the diversity/uncertainty from human labeling.

Moreover, depth data in the RGB-D saliency dataset can be noisy, and a direct fusion of RGB and depth information may overwhelm the network to fit noise. To deal with the noisy depth problem, a depth correction network is proposed as an auxiliary component to produce depth images with rich semantic and geometric information. We also introduce a saliency consensus module to mimic the majority voting mechanism for saliency GT generation.

Our main contributions are summarized as: 1) We propose a conditional probabilistic RGB-D saliency prediction model that can produce diverse saliency predictions instead of a single saliency map;
2) We provide a mechanism via saliency consensus to better model how saliency detection works; 3) We present a depth correction network to decrease noise that is inherent in depth data; 4) Extensive experimental results on six RGB-D saliency detection benchmark  datasets demonstrate the effectiveness of our \ourmodel.


\section{Related Work}
\subsection{RGB-D Saliency Detection}
Depend on how the complementary information between RGB images and depth images is fused, existing RGB-D saliency detection models can be roughly classified into three categories: early-fusion models \cite{qu2017rgbd}, late-fusion models \cite{wang2019adaptive,han2017cnns} and cross-level fusion models \cite{dmra_iccv19,chen2018progressively,chen2019multi,chen2019three,zhao2019Contrast}. 
Qu \etal \cite{qu2017rgbd} proposed an early-fusion model to generate feature for each superpixel of the RGB-D pair, which was then fed to a CNN to produce saliency of each superpixel. Recently, Wang~\etal~\cite{wang2019adaptive} introduced a late-fusion network (\ie AFNet) to fuse predictions from the RGB and depth branch adaptively. In a similar pipeline, Han \etal \cite{han2017cnns} fused the RGB and depth information through fully connected layers.
Chen \etal \cite{chen2019multi} used a multi-scale multi-path network for different modality information fusion. Chen \etal \cite{chen2018progressively} proposed a complementary-aware RGB-D saliency detection model by fusing features from the same stage of each modality with a complementary-aware fusion block.
Chen \etal \cite{chen2019three} presented attention-aware cross-level combination blocks for multi-modality fusion.
Zhao \etal \cite{zhao2019Contrast} integrated a contrast prior to enhance depth cues, and employed a fluid pyramid integration framework to achieve multi-scale cross-modal feature fusion.
To effectively incorporate geometric and semantic information within a recurrent learning framework, Li \etal~\cite{dmra_iccv19} introduced a depth-induced multi-scale RGB-D saliency detection network.

\begin{figure*}[!htp]
   \begin{center}
   {\includegraphics[width=0.85\linewidth]{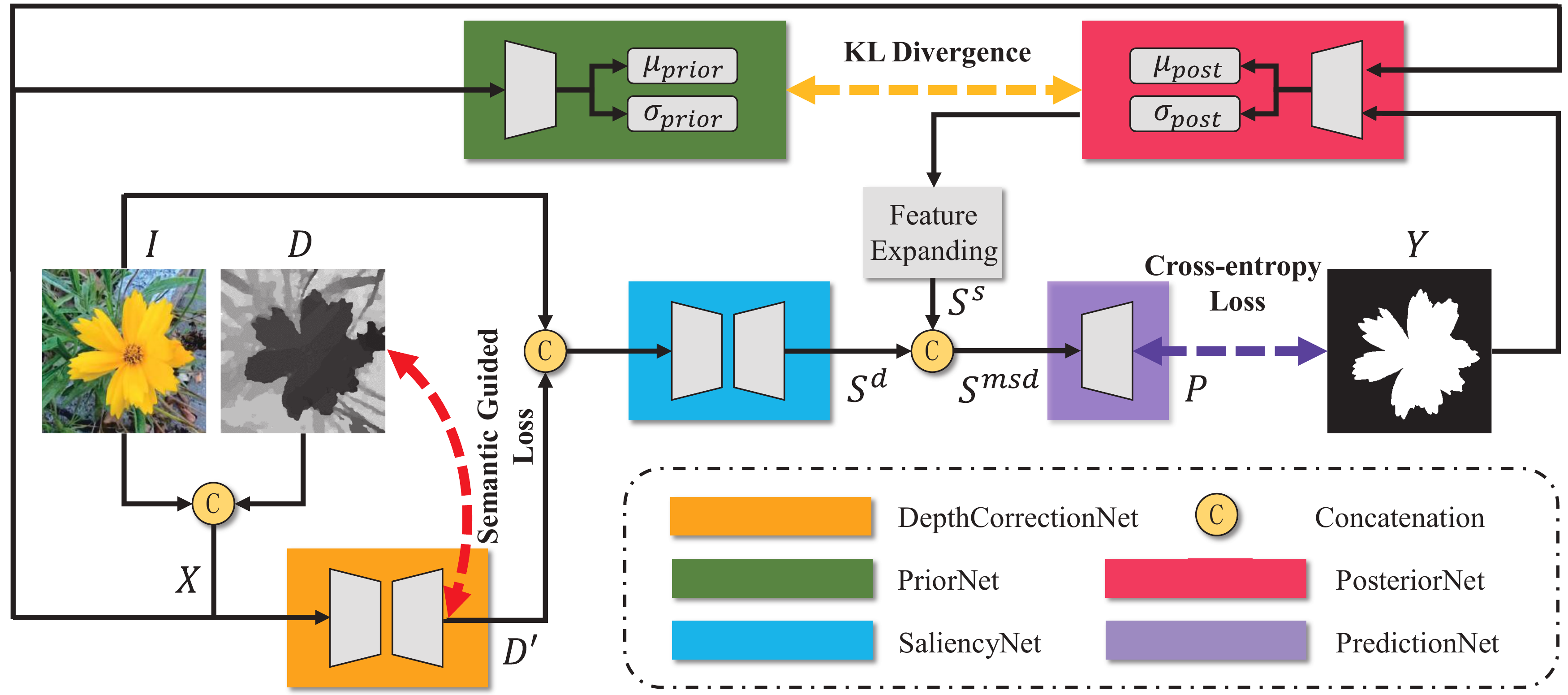}} 
   \end{center}
   \vspace{-6pt}
   \caption{Network training pipeline. Four main modules are included, namely a LatentNet (PriorNet $(\mu_\mathrm{prior},\sigma_\mathrm{prior})$ and PosteriorNet
   $(\mu_\mathrm{post},\sigma_\mathrm{post})$), a SaliencyNet, a DepthCorrectionNet
   and a PredictionNet. The LatentNet maps the RGB-D image pair $X$ (or together with GT $Y$ for the PosteriorNet) to low dimensional Gaussian latent variable $z$.  The DepthCorrectionNet
   refines the raw depth with a semantic guided loss. The SaliencyNet takes the RGB image and the refined depth as input to generate a saliency feature map. The PredictionNet takes both stochastic features and deterministic features to produce a final saliency map. We perform saliency consensus in the testing stage, as shown in Fig. \ref{fig:testing_overview} to generate the final saliency map according to the mechanism of GT saliency map generation.}
   
   

   \label{fig:overview}
   \vspace{-4mm}
\end{figure*}

\begin{figure}[!htp]
   \begin{center}
   {\includegraphics[width=1\linewidth]{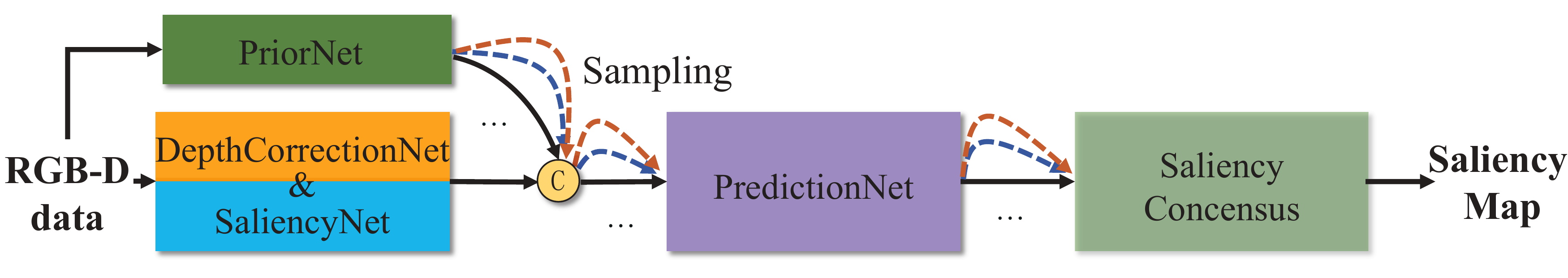}} 
   \end{center}
   \vspace{-7pt}
   \caption{Overview of the proposed framework during testing. We sample the PriorNet multiple times to generate diverse and accurate predictions. The saliency consensus module is then used to obtain the majority voting of the final predictions.}
   \label{fig:testing_overview}
   \vspace{-4mm}
\end{figure}


\subsection{VAE or CVAE based Deep Probabilistic Models}
Ever since the seminal work by Kingma \etal \cite{vae_bayes_kumar} and Rezende \etal \cite{pmlr-v32-rezende14}, variational autoencoder (VAE) and its conditional counterpart CVAE \cite{structure_output} have been widely applied in various computer vision problems. 
To train a VAE, a reconstruction loss and a regularizer are needed to penalize the disagreement of the prior and posterior distribution of the latent representation. 
Instead of defining the prior distribution of the latent representation as a standard Gaussian distribution, CVAE utilizes the input observation to modulate the prior on Gaussian latent variables to generate the output.
In low-level vision, VAE and CVAE have been applied to the tasks such as image background modeling~\cite{SuperVAE_AAAI19}, latent representations with sharp samples~\cite{pixel_vae}, difference of motion modes~\cite{MT-VAE}, medical image segmentation model~\cite{PHiSeg2019}, and modeling inherent ambiguities of an image~\cite{probabilistic_unet}.
Meanwhile, VAE and CVAE have been explored in more complex vision tasks such as uncertain future forecast~\cite{ContrastiveVAE,vae_future}, human motion prediction~\cite{aliakbarian2019learning}, and shape-guided image generation~\cite{Esser_2018_CVPR}. Recently, VAE algorithms have been extened to 3D domain targeting applications such as 3D meshes deformation~\cite{Tan_2018_CVPR}, and point cloud instance segmentation~\cite{Yi_2019_CVPR}. 


To the best of our knowledge, CVAE has not been exploited in saliency detection. Although Li \etal \cite{SuperVAE_AAAI19} adopted VAE in their saliency prediction framework, they used VAE to model the image background, and separated salient objects from the background through the reconstruction residuals. In contrast, we use CVAE to model labeling variants, indicating human uncertainty of labeling. We are the first to employ CVAE in saliency prediction network by considering the human uncertainty in annotation.

\section{Our Model}\label{sec:OurApproach}
In this section, we present our probabilistic RGB-D saliency detection model based on a conditional variational autoencoder, which learns the distribution of saliency maps rather than a single prediction.
Let $\xi = \{X_i,Y_i\}_{i=1}^N$ be the training dataset, where $X_i = \{I_i, D_i\}$ denotes the RGB-D input (consisting of the RGB image $I_i$ and the depth image $D_i$), $Y_i$ denotes the ground truth saliency map. 
The whole pipeline of our model during training and testing are illustrated in Fig.~\ref{fig:overview} and Fig.~\ref{fig:testing_overview}, respectively. 

Our network is composed of five main modules: 1) LatentNet (PriorNet and PosteriorNet) that maps the RGB-D input $X_i$ (for PriorNet) or $X_i$ and $Y_i$ (for PosteriorNet) to the low dimensional latent variables $z_i \in \mathbb{R}^K$ ($K$ is dimension of the latent space); 2) DepthCorrectionNet that takes $I_i$ and $D_i$ as input to generate a refined depth image $D'_i$; 3) SaliencyNet that maps the RGB image $I_i$ and the refined depth image $D'_i$ to saliency feature maps $S_i^d$; 4) PredictionNet that employs stochastic features $S_i^s$ from LatentNet and deterministic features $S_i^d$ from SaliencyNet to produce our saliency map prediction $P_i$; 5) A saliency consensus module in the testing stage that mimics the mechanism of saliency GT generation to evaluate the performance with the provided single GT saliency map $Y_i$. We will introduce each module as follows.

\subsection{Probabilistic RGB-D Saliency Model via CVAE}

The Conditional Variational Autoencoder (CVAE) modulates the prior as a Gaussian distribution with parameters conditioned on the input data $X$. There are three types of variables in the conditional generative model:
conditioning variable $X$ (RGB-D image pair in our setting), latent variable $z$, and output variable $Y$. 
For the latent variable $z$ drawn from the Gaussian distribution $P_\theta(z|X)$, the output variable $Y$ is generated from $P_\omega(Y|X,z)$,
then the posterior of $z$ is formulated as $Q_\phi(z|X,Y)$. The loss of CVAE is defined as:
\begin{equation}
\label{CVAE_equation}
\begin{aligned}
    \mathcal{L}_{\mathrm{CVAE}} = E_{z\sim Q_\phi(z|X,Y)}[-\log P_\omega(Y|X,z)] \\
    + D_{KL}(Q_\phi(z|X,Y)||P_\theta(z|X)),
\end{aligned}
\vspace{-5pt}
\end{equation}
where $P_\omega(Y|X,z)$ is the likelihood of $P(Y)$ given latent variable $z$ and conditioning variable $X$, the Kullback-Leibler Divergence $D_{KL}(Q_\phi(z|X,Y)||P_\theta(z|X))$ works as a regularization loss to reduce the gap between the prior $P_\theta(z|X)$ and the auxiliary posterior $Q_\phi(z|X,Y)$. 
In this way, CVAE aims to model the log likelihood $P(Y)$ under encoding error $D_{KL}(Q_\phi(z|X,Y)||P_\theta(z|X))$.
Following the standard practice in conventional CVAE \cite{structure_output}, we design a CVAE-based RGB-D saliency detection network, and describe each component of our model in the following.

\begin{figure}[t!]
	\centering
    \small
	\begin{overpic}[width=1.0\columnwidth]{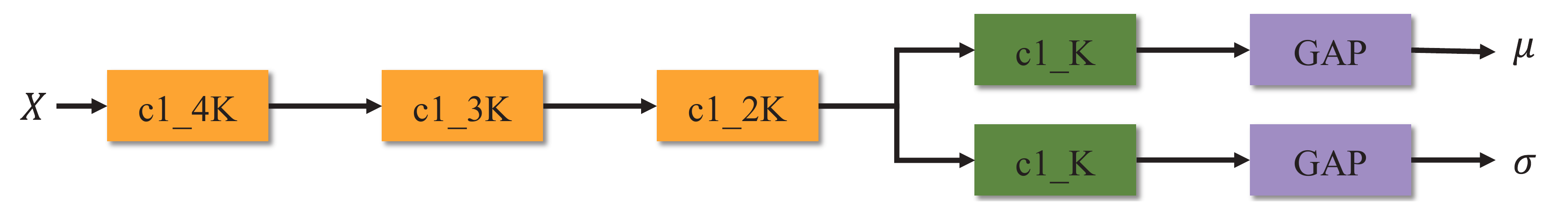}
    \end{overpic}
	\caption{Detailed structure of LatentNet,
   where $K$ is dimension of the latent space, \enquote{c1\_4K} represents a $1\times1$ convolutional layer of output channel size $4K$, \enquote{GAP} is global average pooling.}
    \vspace{-4mm}
    \label{fig:encoder_latent}
\end{figure}

\noindent\textbf{LatentNet:} 
We define $P_\theta(z|X)$ as PriorNet that maps the input RGB-D image pair $X$ to a low-dimensional latent feature space, where $\theta$ is the parameter set of PriorNet. With the same network structure and provided GT saliency map $Y$, we define $Q_\phi(z|X, Y)$ as PosteriorNet, with $\phi$ being the posterior net parameter set. 
In the LatentNet (PriorNet and PosteriorNet), we use five convolutional layers to map the input RGB-D image $X$ (or concatenation of $X$ and $Y$ for the PosteriorNet) to the latent Gaussian variable
$z \sim \mathcal{N}(\mu,\mathrm{diag}(\sigma^2))$,
where $\mu$, $\sigma$ $\in \mathbb{R}^K$, representing the mean and standard deviation of the latent Gaussian variable, as shown in Fig. \ref{fig:encoder_latent}.




Let us define parameter set of PriorNet and PosteriorNet as $(\mu_{\mathrm{prior}}, \sigma_{\mathrm{prior}})$ and $(\mu_{\mathrm{post}}, \sigma_{\mathrm{post}})$ respectively. The KL-Divergence in Eq. \eqref{CVAE_equation} is used to measure the distribution mismatch between the prior net $P_\theta(z|X)$ and posterior net $Q_\phi(z|X,Y)$, or how much information is lost when using $Q_\phi(z|X,Y)$ to represent $P_\theta(z|X)$.
Typical using of CVAE involves multiple versions of ground truth $Y$ \cite{probabilistic_unet} to produce informative $z \in \mathbb{R}^K$, with each position in $z$ represents possible labeling variants or factors that may cause diverse saliency annotations. As we have only one version of GT, directly training with the provided single GT may fail to produce diverse predictions as the network will simply fit the provided annotation $Y$.

\begin{figure}[thp]
	\centering
    \small
	\begin{overpic}[width=1.0\columnwidth]{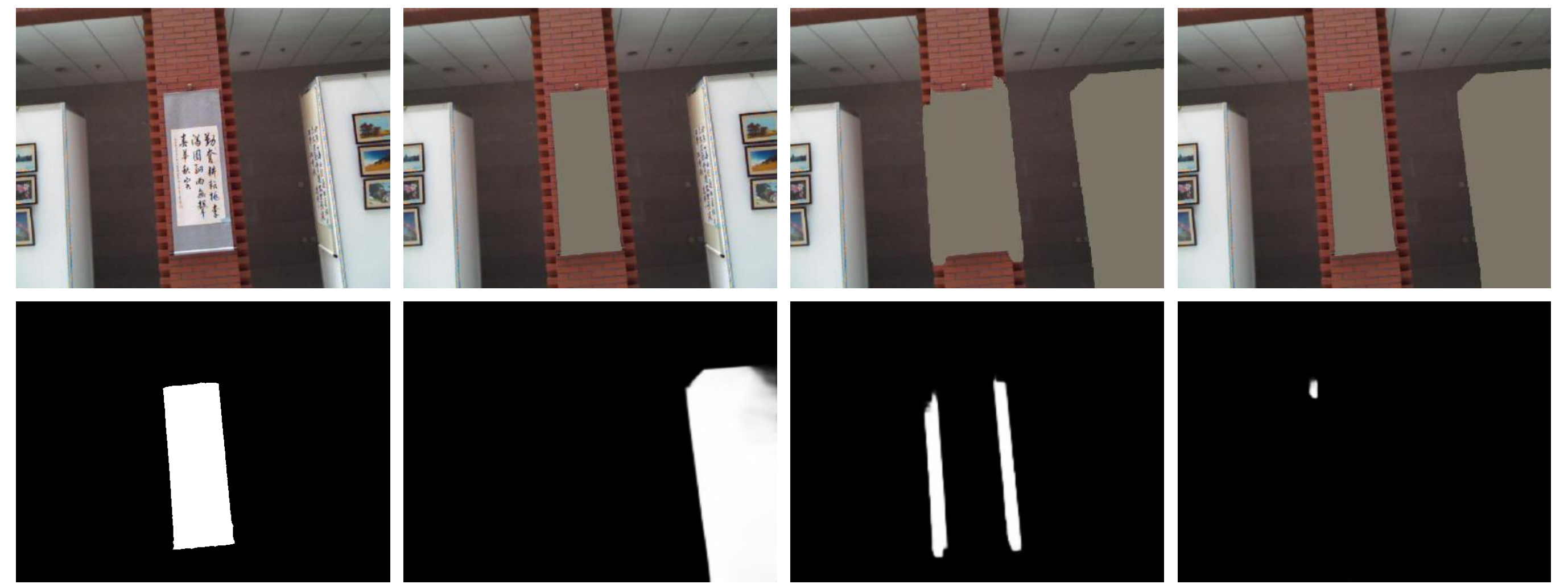}
    \end{overpic}
    \caption{New label generation. The 1$^{st}$ row: we iteratively hide the predicted salient region, where no region is hidden in the first image. The 2$^{nd}$ row: the corresponding GT of the hidden image.}
   \label{fig:iterative_label_generation}
   \vspace{-1mm}
\end{figure}

\noindent\textit{Generate Multiple Predictions:} To produce diverse and accurate predictions, we propose an iterative hiding technique inspired by \cite{hide_and_seek-iccv2017} following the orientation shifting theory \cite{ITTI20001489} to generate more
annotations as shown in Fig. \ref{fig:iterative_label_generation}. We iteratively hide the salient region in the RGB image with mean of the training dataset. The RGB image and its corresponding GT are set as the starting point of the \enquote{new label generation} technique. We first hide the ground truth salient object in the RGB image, and feed the modified image to an existing RGB saliency detection model \cite{BASNet_Sal} to produce a saliency map and treat it as one candidate annotation.
We repeat salient object hiding technique three times for each training image\footnote{We found that usually after three times of hiding, there exists no salient objects in the hidden image.} to obtain four different sets of annotations in total (including the provided GT), and we term this dataset as \enquote{AugedGT}, which is our training dataset.


During training, different annotations (as shown in Fig. \ref{fig:iterative_label_generation}) in $Q_\phi(z|X,Y)$ can force the PriorNet $P_\theta(z|X)$ to encode labeling variants of a given input $X$. As we have already obtained diverse annotations with the proposed hiding technique,
we are expecting the network to produce diverse predictions for images with complicated context. During testing, we can obtain one stochastic feature $S^s$ (input of the \enquote{PredictionNet}) of channel size $K$ each time we sample as shown in Fig. \ref{fig:testing_overview}.


\begin{figure}[tbp]
	\centering
    \small
	\begin{overpic}[width=1.0\columnwidth]{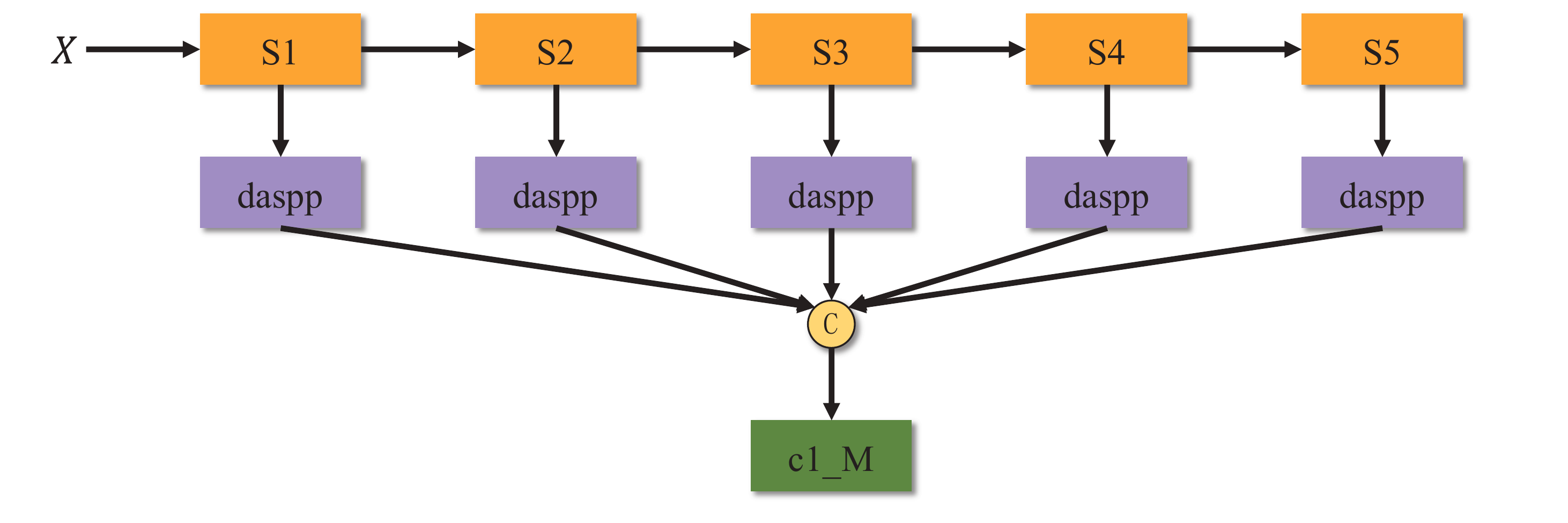}
    \end{overpic}
    \caption{SaliencyNet, where \enquote{S1} represents the first stage of the VGG16 network, \enquote{daspp} is the DenseASPP module \cite{denseaspp}.}
   \label{fig:saliency_feature_net}
   \vspace{-4mm}
\end{figure}

\noindent\textbf{SaliencyNet:}
We design SaliencyNet to produce a deterministic saliency feature map $S^d$ from the input RGB-D data, where the refined depth data comes from the DepthCorrectionNet. We use VGG16 \cite{VGG} as our encoder, and remove layers after the fifth pooling layer. To enlarge the receptive field, we follow DenseASPP~\cite{denseaspp} to obtain feature map with the receptive field of the whole image on each stage of the VGG16 network. We then concatenate those feature maps and feed it to another convolutional layer to obtain
$S^d$.
The detail of the SaliencyNet is illustrated in Fig. \ref{fig:saliency_feature_net}, where \enquote{c1\_M} represents convolutional layer of kernel size $1\times1$, and $M$ is channel size of $S^d$.




\noindent\textit{Feature Expanding:} Statistics ($z\sim\mathcal{N}(\mu,\mathrm{diag}(\sigma^2))$ in particular) from the LatentNet (PriorNet during testing as shown in Fig. \ref{fig:testing_overview} \enquote{Sampling}, or PosteriorNet during training in Fig. \ref{fig:overview}) form the input to 
the Feature Expanding module. Given a pair of $(\mu^k,\sigma^k)$ in each position of the $K$ dimensional vector, we obtain latent vector $z^k=\sigma^k \odot \epsilon + \mu^k$, where $\epsilon \in \mathcal{N}(0,\mathbf{I})$. To fuse with deterministic feature $S^d$, we expand $z^k$ to feature map of the same spatial size as $S^d$ by defining $\epsilon$ as two-dimensional Gaussian noise map. With $k=1,...,K$, we can obtain 
a $K$
(size of the latent space) 
channel stochastic feature $S^s$ representing labeling variants.

\noindent\textbf{PredictionNet:}
The LatentNet produces stochastic features $S^s$ representing labeling variants, while the SaliencyNet outputs deterministic saliency features $S^d$ of input $X$. We propose the PredictionNet, as shown in Fig. \ref{fig:overview} to fuse features from mentioned branches.
A naive concatenation of $S^s$ and $S^d$ may lead the network to learn only from the deterministic features,
thus fail to model labeling variants. Inspired by~\cite{aliakbarian2019learning}, 
we mix $S^s$ and $S^d$ channel-wise; thus, the network cannot distinguish between features of the deterministic branch and the probabilistic branch.
We concatenate $S^d$ and $S^s$ to form a $K+M$ channel feature map $S^{sd}$. We define $K+M$ dimensional variable $r$ (a learnable parameter) representing possible ranking of $1,2,...,K+M$, and then $S^{sd}$ is mixed channel-wisely according to $r$ 
to obtain the mixed feature $S^{msd}$.
Three $1\times1$ convolutional layers with output channel sizes of $K, K/2,1$, are included in the PredictionNet to map $S^{msd}$ to a single channel saliency map $P$. 
During testing, with multiple stochastic features $S^s$, we can obtain multiple predictions by sampling $S^s$ from the LatentNet $\mathcal{N}(\mu_\mathrm{prior},diag(\sigma^2_\mathrm{prior}))$ multiple times.

\subsection{DepthCorrectionNet}
Two main approaches are employed to acquire depth data for RGB-D saliency detection: through depth sensors such as Microsoft Kinect, \eg, DES \cite{cheng2014depth}, and NLPR \cite{peng2014rgbd} datasets; or computing depth from stereo cameras, examples of such datasets are SSB \cite{niu2012leveraging} and NJU2K \cite{NJU2000}. Regardless of the capturing technique, noise is inherent in the depth data. 
We propose a semantic guided depth correction network to produce refined depth information 
as shown in Fig. \ref{fig:overview}, termed as \enquote{DepthCorrectionNet}. The encoder part of the DepthCorrectionNet is the same as the \enquote{SaliencyNet}, while the decoder part is composed of four sequential convolutional layers and bilinear upsampling operation.


We assume that edges of the depth map should be aligned with edges of the RGB image. We adopt the boundary IOU loss \cite{Luo2017CVPR} as a regularizer for DepthCorrectionNet to achieve a refined depth, which is guided by intensity of the RGB image. The full loss for DepthCorrectionNet is defined as:
\begin{equation}
\label{depth_loss}
    \mathcal{L}_{\mathrm{Depth}}=\mathcal{L}_{sl}+\mathcal{L}_{\mathrm{Ioub}},
\vspace{-5pt}
\end{equation}
where $\mathcal{L}_{sl}$ is the smooth $\ell_1$ loss between the refined depth $D'$ and the raw depth $D$, $\mathcal{L}_{ioub}$ is the boundary IOU loss between the refined depth $D'$ and intensity $Ig$ of the RGB image $I$.
Given the predicted depth map $D'$ and intensity of RGB image $Ig$, we follow \cite{Luo2017CVPR} to compute the first-order derivatives of $D'$ and $Ig$. Subsequently, we calculate the magnitude $gD'$ and $gI$ of the gradients of $D'$ and $Ig$, and define the boundary IOU loss as:
\begin{equation}
\label{boundary_iou_loss}
    \mathcal{L}_{\mathrm{Ioub}} = 1-2\frac{|gD' \cap gI|}{|gD'|+|gI|}.
\vspace{-5pt}
\end{equation}

\subsection{Saliency Consensus Module}
Saliency detection is subjective to some extent, and it is common to have multiple annotators to label one image, and the final ground truth saliency map is obtained through majority voting strategy \cite{sip_dataset}.
Although it is well known in the saliency detection community about how the ground truth is acquired; yet, there exists no research on embedding this mechanism into deep saliency frameworks. \textit{Current models define saliency detection as a point estimation problem instead of a distribution estimation problem}. We, instead, use CVAE to obtain the saliency distribution. Next, we embed saliency consensus into our probabilistic framework to compute the majority voting of different predictions in the testing stage as shown in Fig. \ref{fig:testing_overview}.




During testing, we sample PriorNet with fixed $\mu_{\mathrm{prior}}$ and $\sigma_{\mathrm{prior}}$ to obtain a stochastic feature $S^s$. With each $S^s$ and deterministic feature $S^d$ from SaliencyNet, we obtain one version of saliency prediction $P$. To obtain $C$ different predictions $P^1,..., P^C$, we sample PriorNet $C$ times.
We simultaneously feed these multiple predictions to the saliency consensus module to obtain the consensus of predictions.

Given multiple predictions $\{P^c\}_{c=1}^C$, where $P^c \in [0,1]$, we first compute the binary\footnote{As the GT map $Y\in \{0,1\}$, we produce series of binary predictions with each one representing annotation from one saliency annotator.} version $P_b^c$ of the predictions by performing adaptive threshold \cite{borji2015salient} on $P^c$.
For each pixel $(u,v)$, we obtain a $C$ dimensional feature vector $P_{u,v}\in\{0,1\}$. We define $P^{mjv}_b \in \{0,1\}$ as a one-channel saliency map representing majority voting of $P_{u,v}$.
We define an indicator $\mathbf{1}^c(u,v)=\mathbf{1}(P^c_b(u,v)=P^{mjv}_b(u,v))$ representing whether the binary prediction is consistent with the majority voting of the predictions. If $P^c_b(u,v)=P^{mjv}_b(u,v)$, then $\mathbf{1}^c(u,v)=1$. Otherwise, $\mathbf{1}^c(u,v)=0$. We obtain one gray saliency map after saliency consensus as:
\begin{equation}
\label{majority_voting_equation}
{\small
    P^{mjv}_g(u,v)=\frac{\sum_{c=1}^C \mathbf{1}^c(u,v)}{C} \sum_{c=1}^C(P^c_b(u,v)\}\times \mathbf{1}^c(u,v)).
}
\vspace{-5pt}
\end{equation}

\subsection{Objective Function} 
At this stage, our loss function is composed of two parts \ie $\mathcal{L}_{\mathrm{CVAE}}$ and $\mathcal{L}_{\mathrm{Depth}}$. Furthermore, we propose to use the smoothness loss \cite{UnsupeGodard} as a regularizer to achieve edge-aware saliency detection, based on the assumption of inter-class distinction and intra-class similarity.
Following \cite{occlusion_aware}, we define first-order derivatives of the saliency map in the smoothness term as
\begin{equation}
\label{smoothness_loss}
    \mathcal{L}_{\mathrm{Smooth}} = \sum_{u,v} \sum_{d\in{\overrightarrow{x},\overrightarrow{y}}} \Psi(|\partial_d P_{u,v}|e^{-\alpha |\partial_d Ig(u,v)|}),
    \vspace{-5pt}
\end{equation}
where $\Psi$ is defined as $\Psi(s) = \sqrt{s^2+1e^{-6}}$,
$P_{u,v}$ is the predicted saliency map at position $(u,v)$, and $Ig(u,v)$ is the image intensity, $d$ indexes over partial derivative on $\overrightarrow{x}$ and $\overrightarrow{y}$ directions. We set $\alpha=10$ following \cite{occlusion_aware}.

Both the smoothness loss (Eq. \eqref{smoothness_loss}) and the boundary IOU loss (Eq. \eqref{boundary_iou_loss}) need intensity $Ig$. We convert the RGB image $I$ to a gray-scale intensity image $Ig$ as \cite{Saliency_preserving_eccv}:
\begin{equation}
    Ig = 0.2126\times I^{lr} + 0.7152\times I^{lg} + 0.0722\times I^{lb},
    \vspace{-5pt}
\end{equation}
where $I^{lr}$, $I^{lg}$ and $I^{lb}$ represent the color components in the linear color space after Gamma function been removed from the original color space. $I^{lr}$ is achieved via:
\begin{equation}
\label{gamma_extension}
\small
    I^{lr}=\left\{
\begin{aligned}
\frac{I^r}{12.92} & , & I^r\leq 0.04045, \\
\bigg(\frac{I^r+0.055}{1.055}\bigg)^{2.4} & , & I^r> 0.04045.
\end{aligned}
\right.
\vspace{-5pt}
\end{equation}
where $I^r$ is the original red channel of image $I$, and we compute $I^g$ and $I^b$ in the same way as Eq. \eqref{gamma_extension}.

With smoothness loss $\mathcal{L}_{\mathrm{Smooth}}$, depth loss $\mathcal{L}_{\mathrm{Depth}}$ and CVAE loss $\mathcal{L}_{\mathrm{CVAE}}$, our final loss function is defined as:
\begin{equation}
\label{final_loss}
    \mathcal{L}_{\mathrm{sal}} = \mathcal{L}_{\mathrm{CVAE}}+\lambda_1 \mathcal{L}_{\mathrm{Depth}}+\lambda_2 \mathcal{L}_{\mathrm{Smooth}}.
    \vspace{-5pt}
\end{equation}
In our experiments, we set $\lambda_1=\lambda_2=0.3$.

\noindent\textbf{Training details:} We set channel size of $S^d$ as $M=32$, and scale of latent space as $K=8$. We trained our model using Pytorch, and initialized the encoder of SaliencyNet and DepthCorrectionNet with VGG16 parameters pre-trained on ImageNet. Weights of new layers were initialized with $\mathcal{N}(0,0.01)$, and bias was set as constant. We used the Adam method with momentum 0.9 and decreased the learning rate 10\% after each epoch. The base learning rate was initialized as 1e-4. The whole training took 13 hours with training batch size 6 and maximum epoch 30 on a PC with an NVIDIA GeForce RTX GPU. For input image size $352\times352$, the inference time is 0.06s on average.

\section{Experimental Results}

\begin{table*}[t!]
  \centering
  \scriptsize
  \renewcommand{\arraystretch}{1.1}
  \renewcommand{\tabcolsep}{0.9mm}
  \caption{Benchmarking results of ten leading handcrafted feature-based models and eight deep models on six RGBD saliency datasets.  $\uparrow \& \downarrow$ denote larger and smaller is better, respectively. Here, we adopt mean $F_{\beta}$ and mean $E_{\xi}~\cite{Fan2018Enhanced}$.
  }\label{tab:BenchmarkResults}
  \begin{tabular}{rr|cccccccccc|cccccccc|c}
  \hline
  &  &\multicolumn{10}{c|}{Handcrafted Feature based Models}&\multicolumn{8}{c|}{Deep Models}&\multicolumn{1}{c}{} \\
    & Metric &
   LHM  & CDB  & DESM & GP    &
   CDCP & ACSD & LBE & DCMC & MDSF   & SE   & DF   & AFNet& CTMF & MMCI & PCF   & TANet& CPFP & DMRA & \ourmodel \\
   &  & \cite{peng2014rgbd}        & \cite{liang2018stereoscopic}       & \cite{cheng2014depth}          & \cite{ren2015exploiting}              &
        \cite{zhu2017innovative}   & \cite{NJU2000}                 & \cite{feng2016local}  & \cite{cong2016saliency}
        & \cite{song2017depth}   & \cite{guo2016salient} &\cite{qu2017rgbd}       & \cite{wang2019adaptive}& \cite{han2017cnns}    & \cite{chen2019multi}
         & \cite{chen2018progressively}  &\cite{chen2019three}   &   \cite{zhao2019Contrast} & \cite{dmra_iccv19} & Ours \\
  \hline
  \multirow{4}{*}{\textit{NJU2K} \cite{NJU2000}}
    & $S_{\alpha}\uparrow$    & .514 & .632 & .665 & .527 & .669 & .699 & .695 & .686 & .748 & .664 & .763 & .822 & .849 & .858 & .877 & .879 & .878 & .886& \textbf{.897}\\
    & $F_{\beta}\uparrow$     & .328 & .498 & .550 & .357 & .595 & .512 & .606 & .556 & .628 & .583 & .653 & .827 & .779 & .793 & .840 & .841 & .850 & .873 & \textbf{.886}\\
    & $E_{\xi}\uparrow$       & .447 & .572 & .590 & .466 & .706 & .594 & .655 & .619 & .677 & .624 & .700 & .867 & .846 & .851 & .895 & .895 & .910 & .920 & \textbf{.930}\\
    & $\mathcal{M}\downarrow$ & .205 & .199 & .283 & .211 & .180 & .202 & .153 & .172 & .157 & .169 & .140 & .077 & .085 & .079 & .059 & .061 & .053 & .051& \textbf{.043} \\ \hline
  \multirow{4}{*}{\textit{SSB} \cite{niu2012leveraging}}
    & $S_{\alpha}\uparrow$    & .562 & .615 & .642 & .588 & .713 & .692 & .660 & .731 & .728 & .708 & .757 & .825 & .848 & .873 & .875 & .871 & .879 & .835 & \textbf{.903}\\
    & $F_{\beta}\uparrow$     & .378 & .489 & .519 & .405 & .638 & .478 & .501 & .590 & .527 & .611 & .617 & .806 & .758 & .813 & .818 & .828 & .841 & .837 & \textbf{.884}\\
    & $E_{\xi}\uparrow$       & .484 & .561 & .579 &.508  & .751 & .592 & .601 & .655 & .614 & .664 & .692 & .872 & .841 & .873 & .887 & .893 & .911 & .879 & \textbf{.938}\\
    & $\mathcal{M}\downarrow$ & .172 & .166 & .295 & .182 & .149 & .200 & .250 & .148 & .176 & .143 & .141 & .075 & .086 & .068 & .064 & .060 & .051 & .066 & \textbf{.039}\\ \hline
  \multirow{4}{*}{\textit{DES} \cite{cheng2014depth}}
    & $S_{\alpha}\uparrow$    & .578 & .645 & .622 & .636 & .709 & .728 & .703 & .707 & .741 & .741 & .752 & .770 & .863 & .848 & .842 & .858 & .872 & .900 & \textbf{.934}\\
    & $F_{\beta}\uparrow$     & .345 & .502 & .483 & .412 & .585 & .513 & .576 & .542 & .523 & .618 & .604 & .713 & .756 & .735 & .765 & .790 & .824 & .873 & \textbf{.919}\\
    & $E_{\xi}\uparrow$       & .477 & .572 & .566 & .503 & .748 & .613 & .650 & .631 & .621 & .706 & .684 & .809 & .826 & .825 & .838 & .863 & .888 & .933 & \textbf{.967}\\
    & $\mathcal{M}\downarrow$ & .114 & .100 & .299 & .168 & .115 & .169 & .208 & .111 & .122 & .090 & .093 & .068 & .055 & .065 & .049 & .046 & .038 & .030 & \textbf{.019}\\ \hline
  \multirow{4}{*}{\textit{NLPR} \cite{peng2014rgbd}}
    & $S_{\alpha}\uparrow$    & .630 & .632 & .572 & .655 & .727 & .673 & .762 & .724 & .805 & .756 & .806 & .799 & .860 & .856 & .874 & .886 & .888 & .899 & \textbf{.920}\\
    & $F_{\beta}\uparrow$     & .427 & .421 & .430 & .451 & .609 & .429 & .636 & .542 & .649 & .624 & .664 & .755 & .740 & .737 & .802 & .819 & .840 & .865 & \textbf{.891}\\
    & $E_{\xi}\uparrow$       & .560 & .567 & .542 & .571 & .782 & .579 & .719 & .684 & .745 & .742 & .757 & .851 & .840 & .841 & .887 & .902 & .918 & .940 & \textbf{.951}\\
    & $\mathcal{M}\downarrow$ & .108 & .108 & .312 & .146 & .112 & .179 & .081 & .117 & .095 & .091 & .079 & .058 & .056 & .059 & .044 & .041 & .036 & .031 & \textbf{.025}\\ \hline
  \multirow{4}{*}{\textit{LFSD} \cite{li2014saliency}}
    & $S_{\alpha}\uparrow$    & .557 & .520 & .722 & .640 & .717 & .734 & .736 & .753 & .700 & .698 & .791 & .738 & .796 & .787 & .794 & .801 & .828 & .847& \textbf{.864} \\
    & $F_{\beta}\uparrow$     & .396 & .376 & .612 & .519 & .680 & .566 & .612 & .655 & .521 & .640 & .679 & .736 & .756 & .722 & .761 & .771 & .811 & .845 & \textbf{.855}\\
    & $E_{\xi}\uparrow$       & .491 & .465 & .638 & .584 & .754 & .625 & .670 & .682 & .588 & .653 & .725 & .796 & .810 & .775 & .818 & .821 & .863 & .893 & \textbf{.901}\\
    & $\mathcal{M}\downarrow$ & .211 & .218 & .248 & .183 & .167 & .188 & .208 & .155 & .190 & .167 & .138 & .134 & .119 & .132 & .112 & .111 & .088 & .075 & \textbf{.066}\\ \hline
   \multirow{4}{*}{\textit{SIP} \cite{sip_dataset}}
    & $S_{\alpha}\uparrow$    & .511 & .557 & .616 & .588 & .595 & .732 & .727 & .683 & .717 & .628 & .653 & .720 & .716 & .833 & .842 & .835 & .850 & .806 & \textbf{.875}\\
    & $F_{\beta}\uparrow$     & .287 & .341 & .496 & .411 & .482 & .542 & .572 & .500 & .568 & .515 & .465 & .702 & .608 & .771 & .814 & .803 & .821 & .811 & \textbf{.867}\\
    & $E_{\xi}\uparrow$       & .437 & .455 & .564 & .511 & .683 & .614 & .651 & .598 & .645 & .592 & .565 & .793 & .704 & .845 & .878 & .870 & .893 & .844 & \textbf{.914}\\
    & $\mathcal{M}\downarrow$ & .184 & .192 & .298 & .173 & .224 & .172 & .200 & .186 & .167 & .164 & .185 & .118 & .139 & .086 & .071 & .075 & .064 & .085 & \textbf{.051}\\
  \hline
  \end{tabular}
\end{table*}

\begin{figure*}[thp!]
	\centering
    \small
	\begin{overpic}[width=\textwidth]{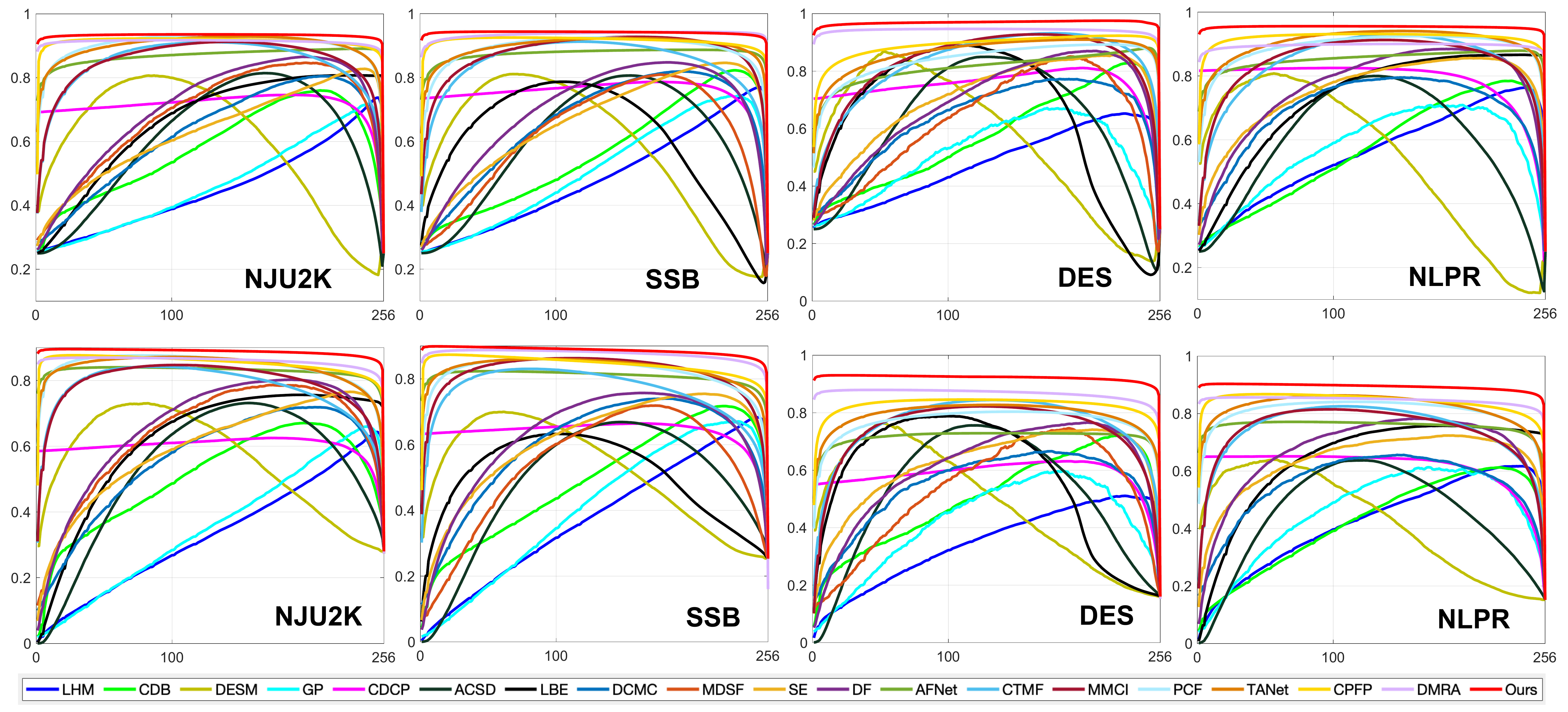}
    \end{overpic}
     \caption{E-measure (1$^{st}$ row) and F-measure (2$^{nd}$ row) curves on four testing datasets.} 
  \label{fig:E_F_measure_show}
  \vspace{-4mm}
\end{figure*}

\subsection{Setup}
\noindent\textbf{Datasets:}
We perform experiments on six datasets including five widely used RGB-D saliency detection datasets (namely NJU2K \cite{NJU2000}, NLPR \cite{peng2014rgbd}, SSB \cite{niu2012leveraging}, LFSD \cite{li2014saliency}, DES \cite{cheng2014depth}) and one newly released dataset (SIP \cite{sip_dataset}). 


\noindent\textbf{Competing Methods:}
We compare our method with 18 algorithms, including ten handcrafted conventional methods and eight deep RGB-D saliency detection models.

\begin{figure*}[t!]
	\centering
    \small
	\begin{overpic}[width=\textwidth]{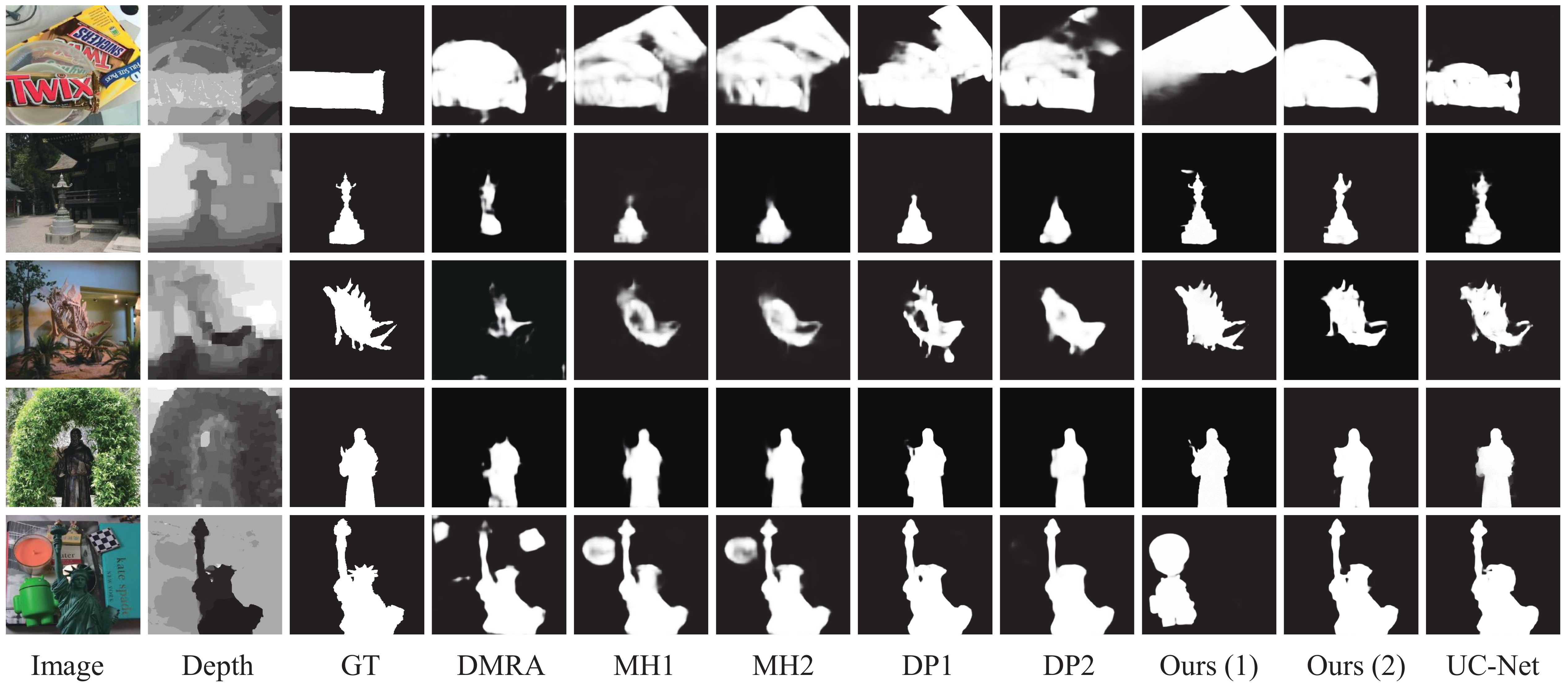}
    \end{overpic}
\caption{\small Comparisons of saliency maps. \enquote{MH1} and \enquote{MH2} are two predictions from M-head. \enquote{DP1} and \enquote{DP2} are predictions of two random MC-dropout during test. \enquote{Ours(1)} and \enquote{Ours(2)} are two predictions sampled from our CVAE based model.
Different from M-head and MC-dropout, which produce consistent predictions for ambiguous images (5$^{th}$ row), \ourmodel~can produce diverse predictions.
}
\vspace{-4mm}
\label{fig:saliency_compare}
\end{figure*}

\noindent\textbf{Evaluation Metrics:}
Four evaluation metrics are used, including two widely used: 1) Mean Absolute Error (MAE $\mathcal{M}$); 2) mean F-measure ($F_{\beta}$) and two recently proposed: 3) Enhanced alignment measure (mean E-measure, $E_{\xi}$) \cite{Fan2018Enhanced} and 4) Structure measure (S-measure, $S_{\alpha}$) \cite{fan2017structure}.
\subsection{Performance Comparison}

\noindent\textbf{Quantitative Comparison:}
We report performance of our method and competing methods in Table \ref{tab:BenchmarkResults}. It shows that our method consistently achieves the best performance on all datasets, especially on SSB \cite{niu2012leveraging} and SIP \cite{sip_dataset}, our method achieves significant 
S-measure, E-measure, and F-measure performance boost and a decrease in MAE by a large margin. 
We show E-measure and F-measure curves of competing methods and ours in Fig. \ref{fig:E_F_measure_show}. 
We observe that our method produces not only stable E-measure and F-measure but also best performance.



\noindent\textbf{Qualitative Comparisons:}
In Fig. \ref{fig:saliency_compare}, we show five images comparing results of our method with one newly released RGB-D saliency detection method (DMRA \cite{dmra_iccv19}), and two widely used methods to produce structured outputs, namely M-head \cite{Rupprecht2016LearningIA} and MC-dropout \cite{kendall2015bayesian} (we will discuss these two methods in detail in the ablation study section). We design both M-head and MC-dropout based structured saliency detection models by replacing CVAE with M-head and MC-dropout respectively. 
Results in Fig. \ref{fig:saliency_compare} show that our method can not only produce high accuracy predictions (compared with DMRA \cite{dmra_iccv19}), but also diverse predictions (compared with M-head based and MC-dropout based models) for images with complex background (image in the first and last rows).


\subsection{Ablation Study}\label{sec:AblationStudy}
We carried out eight experiments (shown in Table \ref{tab:ablation_study}) to thoroughly analyse our framework, including network structure (\enquote{M1}, \enquote{M2}, \enquote{M3}), probabilistic model selection (\enquote{M4}, \enquote{M5}, \enquote{M6}), data source selection (\enquote{M7}) and effectiveness of the new label generation technique (\enquote{M8}).
We make the number bold
when it's better than ours.


\begin{table}[t!]
  \centering
  \scriptsize
  \renewcommand{\arraystretch}{1.1}
  \renewcommand{\tabcolsep}{1.1mm}
  \caption{Ablation study on RGB-D saliency datasets. 
  \vspace{2mm}
  }\label{tab:ablation_study}
  \begin{tabular}{lr|cccccccccccc}
  \hline
    & Metric & \ourmodel &M1  & M2 & M3  & M4     & M5 & M6 & M7& M8& M9 \\
  \hline
  \multirow{4}{*}{\begin{sideways}\textit{NJU2K} \cite{NJU2000}\end{sideways}}
    & $S_{\alpha}\uparrow$    & .897 & .866 & .893 & \textbf{.905} & .871  & .885 & .881& .893& .838& .866  \\
    & $F_{\beta}\uparrow$     & .886  & .858 & \textbf{.887} & .884 & .851  & .878& .878 & .884& .787& .812    \\
    & $E_{\xi}\uparrow$       & .930  & .905 & .930 & .927 & .910  & .923& .927 & \textbf{.932}& .840& .866   \\
    & $\mathcal{M}\downarrow$ & \textbf{.043}  & .060 & .046 & .045 & .059  & .047 & .046 & .044& .084& .075 \\ \hline
 \multirow{4}{*}{\begin{sideways}\textit{SSB} \cite{niu2012leveraging}\end{sideways}}
    & $S_{\alpha}\uparrow$   & \textbf{.903} & .854 & .893 & .900 & .867  & .891 & .893& .898& .855& .872  \\
    & $F_{\beta}\uparrow$    & \textbf{.884}   & .831 & .876 & .868 & .834  & .864 & .876 & .882& .793& .805    \\
    & $E_{\xi}\uparrow$      & \textbf{.938}   & .894 & .911 & .922 & .907  & .921& .931 & .934& .854& .870   \\
    & $\mathcal{M}\downarrow$ & \textbf{.039}  & .060 & .043 & .047 & .057 & .047 & .043 & .040& .073& .068  \\ \hline
\multirow{4}{*}{\begin{sideways}\textit{DES} \cite{cheng2014depth}\end{sideways}}
    & $S_{\alpha}\uparrow$ & \textbf{.934}   & .876 & .896 & .928 & .897  & .911 & .896& .918& .811& .911  \\
    & $F_{\beta}\uparrow$   & \textbf{.919}    & .844 & .868 & .902 & .867  & .897& .868 & .904& .724& .843    \\
    & $E_{\xi}\uparrow$    & \textbf{.967}     & .906 & .928 & .947 & .930  & .945& .928 & .953& .794& .910   \\
    & $\mathcal{M}\downarrow$ & \textbf{.019}  & .035 & .026 & .024 & .033  & .024& .026 & .023& .065& .036  \\ \hline
\multirow{4}{*}{\begin{sideways}\textit{NLPR} \cite{peng2014rgbd}\end{sideways}}
    & $S_{\alpha}\uparrow$ & \textbf{.920}   & .878 & .919 & .918 & .890  & .899 & .910 & .915& .850 & .883  \\
    & $F_{\beta}\uparrow$  & .891     & .846 & \textbf{.897} & .878 & .845  & .875 & .867 & .889 & .759 & .795    \\
    & $E_{\xi}\uparrow$     & .951    & .911 & \textbf{.953} & .941 & .924  & .937 & .933 & .951 & .841 & .883   \\
    & $\mathcal{M}\downarrow$ & .025  & .039 & \textbf{.024} & .029 & .037 & .029  & .028 & .025 & .057 & .045 \\ \hline
    \multirow{4}{*}{\begin{sideways}\textit{LFSD} \cite{li2014saliency}\end{sideways}}
    & $S_{\alpha}\uparrow$ & \textbf{.864}   & .799 & .847 & .862 & .820  & .838 & .847 & .853 & .729& .823 \\
    & $F_{\beta}\uparrow$   & \textbf{.855}    & .791 & .838 & .841  & .802 & .833  & .838 & .848& .661& .779   \\
    & $E_{\xi}\uparrow$    & \textbf{.901}     & .829 & .879 & .885 & .865 & .875  & .879 & .891 & .720& .818  \\
    & $\mathcal{M}\downarrow$ & \textbf{.066}  & .101 & .079 & .075 & .093 & .079  & .079 & .073  & .145& .108\\ \hline
    \multirow{4}{*}{\begin{sideways}\textit{SIP} \cite{sip_dataset}\end{sideways}}
    & $S_{\alpha}\uparrow$ & \textbf{.875}   & .846 & .867 & .870 & .851  & .859 & .867 & .865 & .810& .845 \\
    & $F_{\beta}\uparrow$   & \textbf{.867}    & .837 & .860 & .848 & .821 & .853  & .860 & .855 & .751& .795  \\
    & $E_{\xi}\uparrow$    & \textbf{.914}     & .884 & .908 & .901 & .893 & .905  & .908 & .908 & .816& .852 \\
    & $\mathcal{M}\downarrow$ & \textbf{.051}  & .068 & .056  & .059 & .067 & .057  & .056 & .056& .094& .079 \\ \hline
  \end{tabular}
    \vspace{-4mm}
\end{table} 

\noindent\textbf{Scale of Latent Space:}
We investigate the influence of the scale of the Gaussian latent space $K$ in our network. In this paper, after parameter tuning, we find $K=8$ works best. We show performance with $K=32$ as \enquote{M1}. Performance of \enquote{M1} is worse than our reported results, which indicates that scale of the latent space is an important parameter in our framework. We further carried out more experiments with $K\in[2,12]$, and found relative stable predictions with $K\in[6,10]$. 

\noindent\textbf{Effect of DepthCorrectionNet:}
To illustrate the effectiveness of the proposed DepthCorrectionNet, we remove this branch and feed the concatenation of the RGB image and depth data to the SaliencyNet, shown as \enquote{M2}, which is worse than our method. On DES~\cite{cheng2014depth} dataset, we observe the proposed solution achieves around 4\% improvement on S-measure, E-measure and F-measure, which demonstrates the effectiveness of the depth correction net.

\noindent\textbf{Saliency Consencus Module:}
To mimic the saliency labeling process, we embed a saliency consensus module during test in our framework (as shown in Fig. \ref{fig:testing_overview}) to obtain the majority voting of the multiple predictions. We remove it from our framework and test the network performance by random sample from the latent PriorNet $P_\theta(z|X)$, and performance is shown in \enquote{M3}, which is the best compared with competing methods. While, with the saliency consensus module embedded, we achieve even better performance, which illustrates effectiveness of the saliency consencus module.

\noindent\textbf{VAE $vs.$ CVAE:}
We use CVAE to model labeling variants, and a PosteriorNet is used to estimate parameters for the PriorNet.
To test how our model performs with prior of $z$ as a standard normal distribution,
and the posterior of $z$ as $P_\theta(z|X)$. 
VAE performance is shown as \enquote{M4}, which is comparable with SOTA RGB-D models. With the CVAE \cite{structure_output} based model proposed, we further boost performance of \enquote{M4}, which proves effectiveness of the our solution.

\noindent\textbf{Multi-head $vs.$ CVAE:}
Multi-head models \cite{Rupprecht2016LearningIA} generate multiple predictions with different decoders and a shared encoder, and the loss function is always defined as the closest of the multiple predictions. We remove the LatentNet, and copy the decoder of the SaliencyNet multiple times to achieve multiple predictions (\enquote{M5} in this paper). We report performance in \enquote{M5} as mean of the multiple predictions. \enquote{M5} is better than SOTA models (\eg, DMRA) while there still exists gap between M-head based method (\enquote{M5}) and our CVAE based model (\ourmodel).

\noindent\textbf{Monte-Carlo Dropout $vs.$ CVAE:}
Monte-Carlo Dropout \cite{kendall2015bayesian} uses dropout during the testing stage to introduce stochastic to the network. We follow \cite{kendall2015bayesian} to remove the LatentNet, and use dropout in the encoder and decoder of the SaliencyNet in the testing stage. We repeats five times of random dropout (dropout ratio = 0.1), and report the mean performance as \enquote{M6}. Similar to \enquote{M5}, \enquote{M6} also achieves the best performance comparing with SOTA models (\eg, CPFP and DMRA), while the proposed CVAE based model achieves even better performance.

\noindent\textbf{HHA $vs.$ Depth:}
HHA \cite{gupta2014learning} is a widely used technique that encodes the depth data to three channels: \textbf{h}orizontal disparity, \textbf{h}eight above ground, and the \textbf{a}ngle the pixel’s local surface normal makes with the inferred gravity direction. HHA is widely used in RGB-D related dense prediction models \cite{Du_2019_CVPR, han2017cnns} to obtain better feature representation. To test if HHA also works in our scenario, we replace depth with HHA, and performance is shown in \enquote{M7}. We observe similar performance achieved with HHA instead of the raw depth data.

\noindent\textbf{New Label Generation:}
To produce diverse predictions, we follow \cite{hide_and_seek-iccv2017} and generate diverse annotations for
the training dataset. To illustrate the effectiveness of this strategy, we train with only the SaliencyNet to produce single channel saliency map with RGB-D image as input for simplicity.
\enquote{M8} and \enquote{M9} represent using the provided training dataset and augmented training data respectively. We observe performance improvement of \enquote{M9} compared with \enquote{M8},
which indicates effectiveness of the new label generation technique. 



\section{Conclusion}
\vspace{-5pt}
Inspired by human uncertainty in ground truth (GT) annotation, we proposed the first uncertainty network named \textit{\textbf{UC-Net}} for RGB-D saliency detection based on a conditional variational autoencoder.
Different from existing methods, which generally treat saliency detection as a point estimation problem, we propose to learn the distribution of saliency maps. 
Under our formulation, our model is able to generate multiple labels which have been discarded in the GT annotation generation process through saliency consensus. 
Quantitative and qualitative evaluations on six standard and challenging benchmark datasets demonstrated the superiority of our approach in learning the distribution of saliency maps.
In the future, we would like to extend our approach to other saliency detection problems (\eg, VSOD~\cite{fan2019shifting}, RGB SOD~\cite{fan2018salient,zhao2019egnet}, Co-SOD~\cite{fan2020taking}).
Furthermore, we plan to capture new datasets with multiple human annotations to further model the statistics of human uncertainty in interactive image segmentation~\cite{fClick20CVPR}, camouflaged object detection~\cite{fan2020Camouflage}, \etc.




\vspace{-5pt}
\small{\vspace{.1in}\noindent\textbf{Acknowledgments.}\quad
This research was supported in part by Natural  Science  Foundation  of  China  grants  (61871325, 61420106007, 61671387), the Australia Research Council Centre of Excellence for Robotics Vision (CE140100016), and the National Key Research and Development Program of China under Grant 2018AAA0102803. We thank all reviewers and Area Chairs for their constructive comments.}

{\small
\bibliographystyle{ieee_fullname}
\bibliography{RGBD_Saliency}

\begin{thebibliography}{10}\itemsep=-1pt

\bibitem{ContrastiveVAE}
Abubakar Abid and James~Y. Zou.
\newblock {Contrastive Variational Autoencoder Enhances Salient Features}.
\newblock {\em CoRR}, abs/1902.04601, 2019.

\bibitem{achanta2009frequency}
Radhakrishna Achanta, Sheila Hemami, Francisco Estrada, and Sabine Susstrunk.
\newblock Frequency-tuned salient region detection.
\newblock In {\em IEEE CVPR}, pages 1597--1604, 2009.

\bibitem{PHiSeg2019}
Christian~F. Baumgartner, Kerem~Can Tezcan, Krishna Chaitanya, Andreas~M.
  H{\"{o}}tker, Urs~J. Muehlematter, Khoschy Schawkat, Anton~S. Becker, Olivio
  Donati, and Ender Konukoglu.
\newblock {PHiSeg: Capturing Uncertainty in Medical Image Segmentation}.
\newblock In {\em MICCAI}, pages 119--127, 2019.

\bibitem{borji2015salient}
Ali Borji, Ming-Ming Cheng, Huaizu Jiang, and Jia Li.
\newblock {Salient Object Detection: A Benchmark}.
\newblock {\em IEEE TIP}, 24(12):5706--5722, 2015.

\bibitem{chen2018progressively}
Hao Chen and Youfu Li.
\newblock Progressively complementarity-aware fusion network for {RGB-D Salient
  Object Detection}.
\newblock In {\em IEEE CVPR}, pages 3051--3060, 2018.

\bibitem{chen2019three}
Hao Chen and Youfu Li.
\newblock {Three-stream Attention-aware Network for {RGB-D} Salient Object
  Detection}.
\newblock {\em IEEE TIP}, pages 2825--2835, 2019.

\bibitem{chen2019multi}
Hao Chen, Youfu Li, and Dan Su.
\newblock Multi-modal fusion network with multi-scale multi-path and
  cross-modal interactions for {RGB-D} salient object detection.
\newblock {\em PR}, 86:376--385, 2019.

\bibitem{cheng2014depth}
Yupeng Cheng, Huazhu Fu, Xingxing Wei, Jiangjian Xiao, and Xiaochun Cao.
\newblock Depth enhanced saliency detection method.
\newblock In {\em ACM ICIMCS}, pages 23--27, 2014.

\bibitem{UnsupeGodard}
Gabriel J.~Brostow Clément~Godard, Oisin Mac~Aodha.
\newblock {Unsupervised Monocular Depth Estimation with Left-Right
  Consistency}.
\newblock In {\em IEEE CVPR}, pages 6602--6611, 2017.

\bibitem{cong2016saliency}
Runmin Cong, Jianjun Lei, Changqing Zhang, Qingming Huang, Xiaochun Cao, and
  Chunping Hou.
\newblock Saliency detection for stereoscopic images based on depth confidence
  analysis and multiple cues fusion.
\newblock {\em IEEE SPL}, 23(6):819--823, 2016.

\bibitem{Du_2019_CVPR}
Dapeng Du, Limin Wang, Huiling Wang, Kai Zhao, and Gangshan Wu.
\newblock {Translate-to-Recognize Networks for RGB-D Scene Recognition}.
\newblock In {\em IEEE CVPR}, pages 11836--11845, 2019.

\bibitem{Esser_2018_CVPR}
Patrick Esser, Ekaterina Sutter, and Björn Ommer.
\newblock {A Variational U-Net for Conditional Appearance and Shape
  Generation}.
\newblock In {\em IEEE CVPR}, pages 8857--8865, 2018.

\bibitem{fan2018salient}
Deng-Ping Fan, Ming-Ming Cheng, Jiang-Jiang Liu, Shang-Hua Gao, Qibin Hou, and
  Ali Borji.
\newblock {Salient objects in clutter: Bringing salient object detection to the
  foreground}.
\newblock In {\em ECCV}, pages 186--202, 2018.

\bibitem{fan2017structure}
Deng-Ping Fan, Ming-Ming Cheng, Yun Liu, Tao Li, and Ali Borji.
\newblock {Structure-measure: A new way to evaluate foreground maps}.
\newblock In {\em IEEE ICCV}, pages 4548--4557, 2017.

\bibitem{Fan2018Enhanced}
Deng-Ping Fan, Cheng Gong, Yang Cao, Bo Ren, Ming-Ming Cheng, and Ali Borji.
\newblock {Enhanced-alignment Measure for Binary Foreground Map Evaluation}.
\newblock In {\em IJCAI}, pages 698--704, 2018.

\bibitem{fan2020Camouflage}
Deng-Ping Fan, Ge-Peng Ji, Guolei Sun, Ming-Ming Cheng, Jianbing Shen, and Ling
  Shao.
\newblock {Camouflaged Object Detection}.
\newblock In {\em IEEE CVPR}, 2020.

\bibitem{fan2020taking}
Deng-Ping Fan, Zheng Lin, Ge-Peng Ji, Dingwen Zhang, Huazhu Fu, and Ming-Ming
  Cheng.
\newblock {Taking a Deeper Look at the Co-salient Object Detection}.
\newblock In {\em IEEE CVPR}, 2020.

\bibitem{sip_dataset}
Deng-Ping Fan, Zheng Lin, Zhao Zhang, Menglong Zhu, and Ming-Ming Cheng.
\newblock {Rethinking RGB-D salient object detection: Models, datasets, and
  large-scale benchmarks}.
\newblock {\em IEEE TNNLS}, 2020.

\bibitem{fan2019shifting}
Deng-Ping Fan, Wenguan Wang, Ming-Ming Cheng, and Jianbing Shen.
\newblock Shifting more attention to video salient object detection.
\newblock In {\em IEEE CVPR}, pages 8554--8564, 2019.

\bibitem{feng2016local}
David Feng, Nick Barnes, Shaodi You, and Chris McCarthy.
\newblock Local background enclosure for {RGB-D} salient object detection.
\newblock In {\em IEEE CVPR}, pages 2343--2350, 2016.

\bibitem{Fu2020JLDCF}
Keren~Fu Fu, Deng-Ping Fan, Ge-Peng Ji, and Qijun Zhao.
\newblock {JL-DCF: Joint Learning and Densely-Cooperative Fusion Framework for
  RGB-D Salient Object Detection}.
\newblock In {\em IEEE CVPR}, 2020.

\bibitem{guo2016salient}
Jingfan Guo, Tongwei Ren, and Jia Bei.
\newblock Salient object detection for rgb-d image via saliency evolution.
\newblock In {\em ICME}, pages 1--6, 2016.

\bibitem{gupta2014learning}
Saurabh Gupta, Ross Girshick, Pablo Arbel{\'a}ez, and Jitendra Malik.
\newblock Learning rich features from {RGB-D} images for object detection and
  segmentation.
\newblock In {\em ECCV}, pages 345--360, 2014.

\bibitem{han2017cnns}
Junwei Han, Hao Chen, Nian Liu, Chenggang Yan, and Xuelong Li.
\newblock {CNNs}-based {RGB-D} saliency detection via cross-view transfer and
  multiview fusion.
\newblock {\em IEEE TCYB}, pages 3171--3183, 2018.

\bibitem{pixel_vae}
Faruk Ahmed Adrien Ali Taïga Francesco Visin David Vázquez Aaron C.~Courville
  Ishaan~Gulrajani, Kundan~Kumar.
\newblock {PixelVAE: A Latent Variable Model for Natural Images}.
\newblock In {\em ICLR}, 2016.

\bibitem{ITTI20001489}
Laurent Itti and Christof Koch.
\newblock A saliency-based search mechanism for overt and covert shifts of
  visual attention.
\newblock {\em VR}, 40(10):1489 -- 1506, 2000.

\bibitem{itti_saliency}
Laurent Itti, Christof Koch, and Ernst Niebur.
\newblock A model of saliency-based visual attention for rapid scene analysis.
\newblock {\em IEEE TPAMI}, 20(11):1254--1259, 1998.

\bibitem{NJU2000}
Ran Ju, Ling Ge, Wenjing Geng, Tongwei Ren, and Gangshan Wu.
\newblock Depth saliency based on anisotropic center-surround difference.
\newblock In {\em ICIP}, pages 1115--1119, 2014.

\bibitem{F3Net_aaai2020}
Shuhui~Wang Jun~Wei and Qingming Huang.
\newblock {F3Net: Fusion, Feedback and Focus for Salient Object Detection}.
\newblock In {\em AAAI}, 2020.

\bibitem{kendall2015bayesian}
Alex Kendall, Vijay Badrinarayanan, , and Roberto Cipolla.
\newblock {Bayesian SegNet: Model Uncertainty in Deep Convolutional
  Encoder-Decoder Architectures for Scene Understanding}.
\newblock In {\em BMVC}, 2017.

\bibitem{vae_bayes_kumar}
Diederik~P {Kingma} and Max {Welling}.
\newblock {Auto-Encoding Variational Bayes}.
\newblock In {\em ICLR}, 2013.

\bibitem{probabilistic_unet}
Simon Kohl, Bernardino Romera-Paredes, Clemens Meyer, Jeffrey De~Fauw,
  Joseph~R. Ledsam, Klaus Maier-Hein, S.~M.~Ali Eslami, Danilo Jimenez~Rezende,
  and Olaf Ronneberger.
\newblock {A Probabilistic U-Net for Segmentation of Ambiguous Images}.
\newblock In {\em NeurIPS}, pages 6965--6975, 2018.

\bibitem{scanpath}
Olivier Le~Meur and Thierry Baccino.
\newblock Methods for comparing scanpaths and saliency maps: strengths and
  weaknesses.
\newblock {\em Behavior Research Methods}, 45(1):251--266, 2013.

\bibitem{SuperVAE_AAAI19}
Bo Li, Zhengxing Sun, and Yuqi Guo.
\newblock {SuperVAE: Superpixelwise Variational Autoencoder for Salient Object
  Detection}.
\newblock In {\em AAAI}, pages 8569--8576, 2019.

\bibitem{li2014saliency}
Nianyi Li, Jinwei Ye, Yu Ji, Haibin Ling, and Jingyi Yu.
\newblock Saliency detection on light field.
\newblock In {\em IEEE CVPR}, pages 2806--2813, 2014.

\bibitem{liang2018stereoscopic}
Fangfang Liang, Lijuan Duan, Wei Ma, Yuanhua Qiao, Zhi Cai, and Laiyun Qing.
\newblock Stereoscopic saliency model using contrast and
  depth-guided-background prior.
\newblock {\em Neurocomputing}, 275:2227--2238, 2018.

\bibitem{fClick20CVPR}
Zheng Lin, Zhao Zhang, Lin-Zhuo Chen, Ming-Ming Cheng, and Shao-Ping Lu.
\newblock {Interactive Image Segmentation with First Click Attention}.
\newblock In {\em IEEE CVPR}, 2020.

\bibitem{Liu_2019_ICCV}
Yi Liu, Qiang Zhang, Dingwen Zhang, and Jungong Han.
\newblock {Employing Deep Part-Object Relationships for Salient Object
  Detection}.
\newblock In {\em IEEE ICCV}, 2019.

\bibitem{Luo2017CVPR}
Zhiming Luo, Akshaya Mishra, Andrew Achkar, Justin Eichel, Shaozi Li, and
  Pierre-Marc Jodoin.
\newblock {Non-Local Deep Features for Salient Object Detection}.
\newblock In {\em IEEE CVPR}, 2017.

\bibitem{niu2012leveraging}
Yuzhen Niu, Yujie Geng, Xueqing Li, and Feng Liu.
\newblock Leveraging stereopsis for saliency analysis.
\newblock In {\em IEEE CVPR}, pages 454--461, 2012.

\bibitem{peng2014rgbd}
Houwen Peng, Bing Li, Weihua Xiong, Weiming Hu, and Rongrong Ji.
\newblock Rgbd salient object detection: a benchmark and algorithms.
\newblock In {\em ECCV}, pages 92--109, 2014.

\bibitem{BASNet_Sal}
Xuebin Qin, Zichen Zhang, Chenyang Huang, Chao Gao, Masood Dehghan, and Martin
  Jagersand.
\newblock {BASNet: Boundary-Aware Salient Object Detection}.
\newblock In {\em IEEE CVPR}, 2019.

\bibitem{qu2017rgbd}
Liangqiong Qu, Shengfeng He, Jiawei Zhang, Jiandong Tian, Yandong Tang, and
  Qingxiong Yang.
\newblock {RGBD} salient object detection via deep fusion.
\newblock {\em IEEE TIP}, 26(5):2274--2285, 2017.

\bibitem{ren2015exploiting}
Jianqiang Ren, Xiaojin Gong, Lu Yu, Wenhui Zhou, and Michael Ying~Yang.
\newblock {Exploiting Global Priors for RGB-D Saliency Detection}.
\newblock In {\em IEEE CVPRW}, pages 25--32, 2015.

\bibitem{pmlr-v32-rezende14}
Danilo~Jimenez Rezende, Shakir Mohamed, and Daan Wierstra.
\newblock {Stochastic Backpropagation and Approximate Inference in Deep
  Generative Models}.
\newblock In {\em ICML}, pages 1278--1286, 2014.

\bibitem{Rupprecht2016LearningIA}
Christian Rupprecht, Iro Laina, Maximilian Baust, Federico Tombari, Gregory~D.
  Hager, and Nassir Navab.
\newblock {Learning in an Uncertain World: Representing Ambiguity Through
  Multiple Hypotheses}.
\newblock In {\em IEEE ICCV}, pages 3611--3620, 2017.

\bibitem{aliakbarian2019learning}
Mohammad {Sadegh Aliakbarian}, Fatemeh {Sadat Saleh}, Mathieu {Salzmann}, Lars
  {Petersson}, Stephen {Gould}, and Amirhossein {Habibian}.
\newblock {Learning Variations in Human Motion via Mix-and-Match Perturbation}.
\newblock {\em arXiv e-prints}, page arXiv:1908.00733, 2019.

\bibitem{VGG}
Karen Simonyan and Andrew Zisserman.
\newblock {Very Deep Convolutional Networks for Large-Scale Image Recognition}.
\newblock In {\em ICLR}, 2014.

\bibitem{hide_and_seek-iccv2017}
Krishna~Kumar Singh and Yong~Jae Lee.
\newblock {Hide-and-Seek: Forcing a Network to be Meticulous for
  Weakly-supervised Object and Action Localization}.
\newblock In {\em IEEE ICCV}, 2017.

\bibitem{structure_output}
Kihyuk Sohn, Honglak Lee, and Xinchen Yan.
\newblock {Learning Structured Output Representation using Deep Conditional
  Generative Models}.
\newblock In {\em NeurIPS}, pages 3483--3491, 2015.

\bibitem{song2017depth}
Hangke Song, Zhi Liu, Huan Du, Guangling Sun, Olivier Le~Meur, and Tongwei Ren.
\newblock Depth-aware salient object detection and segmentation via multiscale
  discriminative saliency fusion and bootstrap learning.
\newblock {\em IEEE TIP}, 26(9):4204--4216, 2017.

\bibitem{Tan_2018_CVPR}
Qingyang Tan, Lin Gao, Yu-Kun Lai, and Shihong Xia.
\newblock {Variational Autoencoders for Deforming 3D Mesh Models}.
\newblock In {\em IEEE CVPR}, 2018.

\bibitem{vae_future}
Jacob Walker, Carl Doersch, Harikrishna Mulam, and Martial Hebert.
\newblock {An Uncertain Future: Forecasting from Static Images Using
  Variational Autoencoders}.
\newblock In {\em ECCVW}, pages 835--851, 2016.

\bibitem{wang2019adaptive}
Ningning Wang and Xiaojin Gong.
\newblock {Adaptive Fusion for RGB-D Salient Object Detection}.
\newblock {\em IEEE Access}, 7:55277--55284, 2019.

\bibitem{Iter_Coop_CVPR}
Wenguan Wang, Jianbing Shen, Ming-Ming Cheng, and Ling Shao.
\newblock {An Iterative and Cooperative Top-Down and Bottom-Up Inference
  Network for Salient Object Detection}.
\newblock In {\em IEEE CVPR}, 2019.

\bibitem{occlusion_aware}
Yang Wang, Yi Yang, Zhenheng Yang, Liang Zhao, Peng Wang, and Wei Xu.
\newblock {Occlusion Aware Unsupervised Learning of Optical Flow}.
\newblock In {\em IEEE CVPR}, 2018.

\bibitem{MT-VAE}
Rastogi Akash Villegas Ruben Sunkavalli Kalyan Shechtman Eli Hadap Sunil Yumer
  Ersin Lee~Honglak Yan, Xinchen.
\newblock {MT-VAE: Learning Motion Transformations to Generate Multimodal Human
  Dynamics}.
\newblock In {\em ECCV}, pages 276--293, 2018.

\bibitem{denseaspp}
Maoke Yang, Kun Yu, Chi Zhang, Zhiwei Li, and Kuiyuan Yang.
\newblock {DenseASPP for Semantic Segmentation in Street Scenes}.
\newblock In {\em IEEE CVPR}, pages 3684--3692, 2018.

\bibitem{Yi_2019_CVPR}
Li Yi, Wang Zhao, He Wang, Minhyuk Sung, and Leonidas~J. Guibas.
\newblock {GSPN: Generative Shape Proposal Network for 3D Instance Segmentation
  in Point Cloud}.
\newblock In {\em IEEE CVPR}, 2019.

\bibitem{Saliency_preserving_eccv}
Shivanthan A.~C. Yohanandan, Adrian~G. Dyer, Dacheng Tao, and Andy Song.
\newblock {Saliency Preservation in Low-Resolution Grayscale Images}.
\newblock In {\em ECCV}, 2018.

\bibitem{dmra_iccv19}
Jingjing Li Miao Zhang Huchuan~Lu Yongri~Piao, Wei~Ji.
\newblock {Depth-induced Multi-scale Recurrent Attention Network for Saliency
  Detection}.
\newblock In {\em IEEE ICCV}, 2019.

\bibitem{jing2020weakly}
Jing Zhang, Xin Yu, Aixuan Li, Peipei Song, Bowen Liu, and Yuchao Dai.
\newblock {Weakly-Supervised Salient Object Detection via Scribble
  Annotations}.
\newblock In {\em IEEE CVPR}, 2020.

\bibitem{Zhang_2018_CVPR}
Jing Zhang, Tong Zhang, Yuchao Dai, Mehrtash Harandi, and Richard Hartley.
\newblock {Deep Unsupervised Saliency Detection: A Multiple Noisy Labeling
  Perspective}.
\newblock In {\em IEEE CVPR}, pages 9029--9038, 2018.

\bibitem{zhao2019Contrast}
Jia-Xing Zhao, Yang Cao, Deng-Ping Fan, Ming-Ming Cheng, Xuan-Yi Li, and Le
  Zhang.
\newblock {Contrast Prior and Fluid Pyramid Integration for RGBD Salient Object
  Detection}.
\newblock In {\em IEEE CVPR}, 2019.

\bibitem{zhao2019egnet}
Jia-Xing Zhao, Jiang-Jiang Liu, Deng-Ping Fan, Yang Cao, Jufeng Yang, and
  Ming-Ming Cheng.
\newblock {EGNet: Edge guidance network for salient object detection}.
\newblock In {\em IEEE ICCV}, pages 8779--8788, 2019.

\bibitem{zhu2017innovative}
Chunbiao Zhu, Ge Li, Wenmin Wang, and Ronggang Wang.
\newblock An innovative salient object detection using center-dark channel
  prior.
\newblock In {\em IEEE ICCVW}, 2017.

\end{thebibliography}
}

\end{document}